\documentclass[conference]{IEEEtran}
\usepackage{times}

\usepackage{etoolbox}
\makeatletter
\patchcmd{\@makecaption}
  {\scshape}
  {}
  {}
  {}
\makeatother
\usepackage[
singlelinecheck=false %
]{caption}

\usepackage[numbers]{natbib}
\usepackage{multicol}
\usepackage[bookmarks=true]{hyperref}
\usepackage[dvipsnames]{xcolor}
\usepackage{amsmath,amssymb,amsfonts}
\usepackage{graphicx}
\usepackage{tabularx}
\usepackage{booktabs}

\begin{document}

\title{Synergies Between Affordance and Geometry: \\ 6-DoF Grasp Detection via Implicit Representations}

\author{
    \IEEEauthorblockN{Zhenyu Jiang\IEEEauthorrefmark{2}, Yifeng Zhu\IEEEauthorrefmark{2}, Maxwell Svetlik\IEEEauthorrefmark{2}, Kuan Fang\IEEEauthorrefmark{3}, Yuke Zhu\IEEEauthorrefmark{2}}
    \IEEEauthorblockA{\IEEEauthorrefmark{2}The University of Texas at Austin, \IEEEauthorrefmark{3}Stanford University}
    
}

\maketitle

\begin{abstract}
Grasp detection in clutter requires the robot to reason about the 3D scene from incomplete and noisy perception.
In this work, we draw insight that 3D reconstruction and grasp learning are two intimately connected tasks, both of which require a fine-grained understanding of local geometry details. We thus propose to utilize the synergies between grasp affordance and 3D reconstruction through multi-task learning of a shared representation. Our model takes advantage of deep implicit functions, a continuous and memory-efficient representation, to enable differentiable training of both tasks. We train the model on self-supervised grasp trials data in simulation. Evaluation is conducted on a clutter removal task, where the robot clears cluttered objects by grasping them one at a time. The experimental results in simulation and on the real robot have demonstrated that the use of implicit neural representations and joint learning of grasp affordance and 3D reconstruction have led to state-of-the-art grasping results. Our method outperforms baselines by over 10\% in terms of grasp success rate. Additional results and videos can be found at \href{https://sites.google.com/view/rpl-giga2021}{\textcolor{Blue}{\url{https://sites.google.com/view/rpl-giga2021}}}
\end{abstract}

\IEEEpeerreviewmaketitle

\section{Introduction}

Generating robust grasps from raw perception is an essential task for robots to physically interact with objects in unstructured environments. This task demands the robots to reason about the geometry and physical properties of objects from partially observed visual data, infer a proper grasp pose (3D position and 3D orientation), and move the gripper to the desired grasp configuration for execution. Here we consider the problem of 6-DoF grasp detection in clutter from 3D point cloud of the robot's on-board depth camera. Our goal is to predict a set of candidate grasps on a clutter of objects from partial point cloud for grasping and decluttering.

Robot grasping is a long-standing challenge with decades of research. Pioneer work~\cite{ferrari1992planning,rodriguez2012caging} has cast it as a \emph{geometry-centric} task, typically assuming access to the full 3D model of the objects. Grasps are thereby generated through optimization on analytical models of constraints derived from geometry and physics. In practice, the requirement of ground-truth models has impeded their applicability in unstructured scenes. One remedy is to integrate these model-driven methods with a 3D reconstruction pipeline that builds the object models from perception as the precursor step~\cite{bohg2011mind,varley2017shape,lundell2019robust}. However, it demands solving a full-fledged 3D reconstruction problem, an open challenge in computer vision. Motivated by the new development of machine learning, in particular deep learning, recent work on grasping has shifted focus towards a \emph{data-driven} paradigm~\cite{bohg2013data,kalashnikov2018qt,mahler2017dex}, where deep networks are trained end-to-end on large-scale grasping datasets, either through manual labeling~\cite{jiang2011efficient} or self-exploration~\cite{kalashnikov2018qt,pinto2016supersizing}. Data-driven methods have enabled direct grasp prediction from noisy perception. However, end-to-end deep learning for grasping often suffers from limited generalization within the training domains.

\begin{figure}
    \captionsetup{font=footnotesize}
    \captionsetup{belowskip=-10pt}
    \centering
    \includegraphics[width=\linewidth]{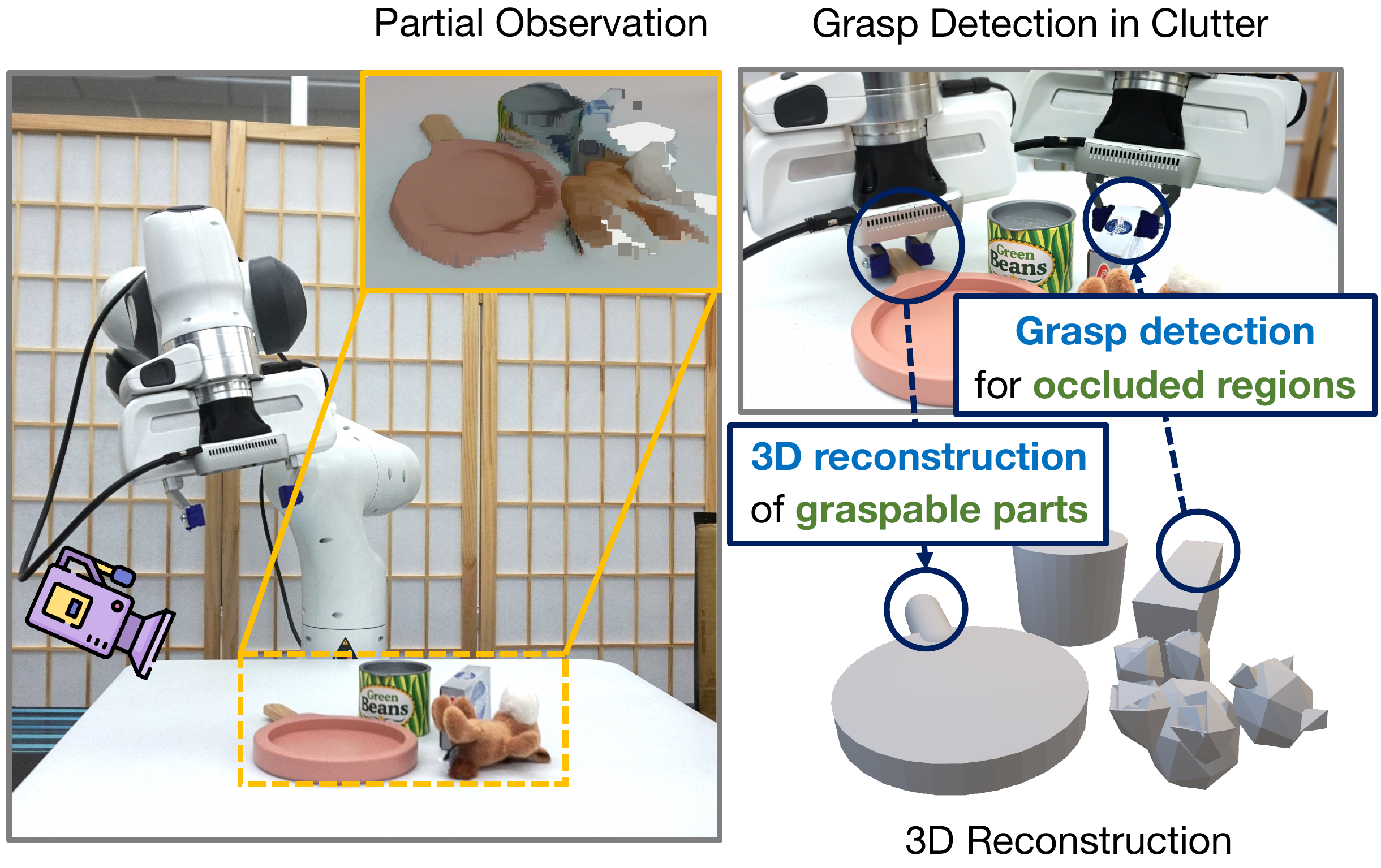}
    \caption{We harness the synergies between affordance and geometry for 6-DoF grasp detection in clutter. Our model jointly learns grasp affordance prediction and 3D reconstruction. Supervision from reconstruction facilitates our model to learn geometrically-aware features for accurate grasps in occluded regions from partial observation. Supervision from grasp, in turn, produce better 3D reconstruction in graspable regions.}
    \label{fig:pull}
\end{figure}

Inspired by the two threads of research on geometry-centric and data-driven approaches to grasping, we investigate the synergistic relations between geometry reasoning and grasp learning. Our key intuition is that \emph{a learned representation capable of reconstructing the 3D scene encodes relevant geometry information for predicting grasp points and vice versa}. In this work, we develop a unified learning framework that enables a shared scene representation for both tasks of grasp prediction and 3D reconstruction, where grasps are represented by a landscape of grasp affordance over the scene. By grasp affordance, we refer to the likelihood of grasp success and the corresponding grasp parameters at each location. %

The primary challenge here is to develop a shared representation that effectively encodes 3D geometry and grasp affordance information. Recent work from the 3D vision and graphics communities has shed light on the merits of implicit representations for geometry reasoning tasks~\cite{mescheder2019occupancy,park2019deepsdf,chen2019learning,peng2020convolutional}. Deep implicit functions define a scene through continuous and differentiable representations parameterized by a deep network. The network maps each spatial location to a corresponding local feature, which can further be decoded to geometry quantities, such as occupancy~\cite{mescheder2019occupancy}, signed distance functions~\cite{park2019deepsdf}, probability density, and emitted color~\cite{liu2020neural, mildenhall2020nerf}. Implicit neural representations have demonstrated state-of-the-art results in a variety of 3D reconstruction tasks~\cite{mescheder2019occupancy,xu2019disn,chibane2020implicit}, due to their ability to represent smooth surfaces in high resolution. On top of the strengths of encoding 3D geometry, the implicit representations are desirable for our problem for two additional reasons: 1) They are differentiable and thus amenable to gradient-based multi-task training of 3D reconstruction and grasp prediction, enabling to learn a shared representation for both tasks; and 2) The representations parameterized by deep networks can adaptively allocate computational budgets to regions of importance. Hence, implicit neural representations can flexibly encode action-related information for the parts of the scene where grasps are likely to succeed.

To this end, we introduce our model: Grasp detection via Implicit Geometry and Affordance (GIGA). We develop a structured implicit neural representation for 6-DoF grasp detection. Our method extracts structured feature grids from the Truncated Signed Distance Function (TSDF) voxel grid fused from the input depth image. A local feature can be computed from the feature grids given a query 3D coordinate. This local feature is used by the implicit functions for estimating the grasp affordance (in the form of grasp quality, grasp orientation, and gripper width of a parallel jaw) and the 3D geometry (in the form of binary occupancy) at the query location. The model is jointly trained in simulation with known 3D geometry and self-supervised grasp trials. With multi-task supervision of affordance and geometry, our model takes advantage of the synergies between them for more robust grasp detection.

We conduct experiments on a clutter removal task~\cite{breyer2020volumetric} in simulation and on physical hardware. In the experiments, multiple objects are piled or packed in clutter, and the robot is tasked to remove the objects by grasping them one at a time. In each round, the robot receives a single-view depth image from the on-board depth camera and predicts a 6-DoF grasp configuration. The ability to reconstruct the 3D scene from a single view enables GIGA to achieve grasp performance on par with multi-view input as employed in prior work~\cite{breyer2020volumetric}. Empirical results have confirmed the benefits of implicit neural representations and the exploitation of the synergies between affordance and geometry. Our model achieved 87.9\% and 69.2\% grasp success rates on the packed and pile scenes, outperforming 74.5\% and 60.7\% reported by the state-of-the-art VGN model. We provide qualitative visualizations of the learned grasp affordance landscape over the entire scene, showing that our implicit representations have encoded scene-level context information for collision and occlusion reasoning. Meanwhile, grasp prediction guides the learned implicit representations to produce better reconstruction to the graspable regions of the scene.

We summarize the main contributions of our work below:

\begin{itemize}
    \item We exploit the synergistic relationships between grasp affordance prediction and 3D reconstruction for 6-DoF grasp detection in clutter.
    \item We introduce structured implicit neural representations that effectively encode 3D scenes and jointly train the representations with simulated self-supervision.
    \item We demonstrate significant improvements of our model over the state-of-the-art in the clutter removal task in simulation and on the real robot.
\end{itemize}

\section{Related Work}

\subsection{Learning Grasp Detection}

Grasping has been studied for decades. Pioneer work has developed analytical methods based on the object models~\cite{ding2000computing,liu1999qualitative,mirtich1994easily,ponce1993characterizing,sahbani2012overview,zhu2004planning}. However, complete models of object geometry and physics are usually unavailable in real-world applications. To bridge these analytical methods with raw perception, prior works have resorted to various 3D reconstruction methods, such as by fitting known CAD models to partial observations~\cite{klank2009real, kragic2001real} or completing full 3D model from partial observations based on symmetry analysis~\cite{bohg2011mind,quispe2015exploiting}. 
In recent years, learning methods, especially deep learning models, have gained increased attention for the grasping problem~\cite{bohg2011mind,james2019sim,kalashnikov2018qt,morrison2018closing,mousavian20196,fang2020learning}. Dex-Net~\cite{mahler2017dex,mahler2019learning} introduced a two-stage pipeline for top-down antipodal grasping. It first samples candidate 4-DoF grasps. The grasp quality of each grasp candidate is then assessed by a convolutional neural network. 6-DoF GraspNet~\cite{mousavian20196} extends grasp generation to the $SE(3)$ space with a variational autoencoder for grasp proposals on a singulated object. GPD~\cite{gualtieri2016high,ten2017grasp} and PointGPD~\cite{liang2019pointnetgpd} tackled  6-DoF grasp detection in clutter with a two-stage grasp pipeline. %
VGN~\cite{breyer2020volumetric} predicts 6-DoF grasps in clutter with a one-stage pipeline from input depth images. There is also a line of works that estimate affordance of an object or a scene first and then detect grasps based on estimated affordance~\cite{pohl2020affordance,kokic2017affordance,yen2020learning}. In most of the prior works, deep networks are trained end-to-end with only grasp supervision. In contrast, our method learns a structured neural representation jointly with self-supervised geometry and grasp supervisions.

\subsection{Geometry-Aware Grasping}

The intimate connection between grasp detection and geometry reasoning has inspired a line of work on geometry-aware grasping. Bohg and Kragic~\cite{bohg2010learning} use a descriptor based on shape context with a non-linear classification algorithm to predict grasps. DGGN~\cite{yan2018learning} regularized grasps through 3D geometry prediction. It learns to predict a voxel occupancy grid from partial observations and evaluates grasp quality from feature of the reconstructed grid. Most relevant to us is PointSDF \cite{van2020learning}, which learns 3D reconstruction via implicit functions and shares the learned geometry features with the grasp evaluation network. It used a global feature for the shared shape representation, and the quantitative results did not show that geometry-aware features improve grasp performances. On the contrary, our work demonstrates that through the use of structured feature grids and implicit functions, GIGA improves grasp detection and focuses more on the 3D reconstruction of graspable regions via joint training of both tasks.

\subsection{Implicit Neural Representations}

Our method is inspired by recent work from the vision and graphics communities on representing 3D objects and scenes with deep implicit functions~\cite{chen2019learning,mescheder2019occupancy,park2019deepsdf}. Rather than explicit  3D representations such as voxels, point clouds, or meshes, these works have used the isosurface of an implicit function to represent the surface of a shape. By parametrizing these implicit functions with deep networks, they are capable of representing complex shapes smoothly and continuously in high resolution. 
The most common architecture for deep implicit functions is multi-layer perceptions (MLP), which encode the geometry information of the entire scene into the model parameters of the MLP. However, they have difficulty in preserving the fine-grained geometric details of local regions. To mitigate this problem, hybrid representations have been introduced to combine feature grid structures and neural representations~\cite{liu2020neural,peng2020convolutional}. These structured representations are appealing for the grasping problem, which requires geometry reasoning in local object parts. Our model extends the architecture of convolutional occupancy network~\cite{peng2020convolutional}, the state-of-the-art scene reconstruction model, as the backbone for joint learning of affordance and geometry.

\section{Preliminaries}

\begin{figure*}[t]
    \captionsetup{font=footnotesize}
    \centering
    \includegraphics[width=\linewidth]{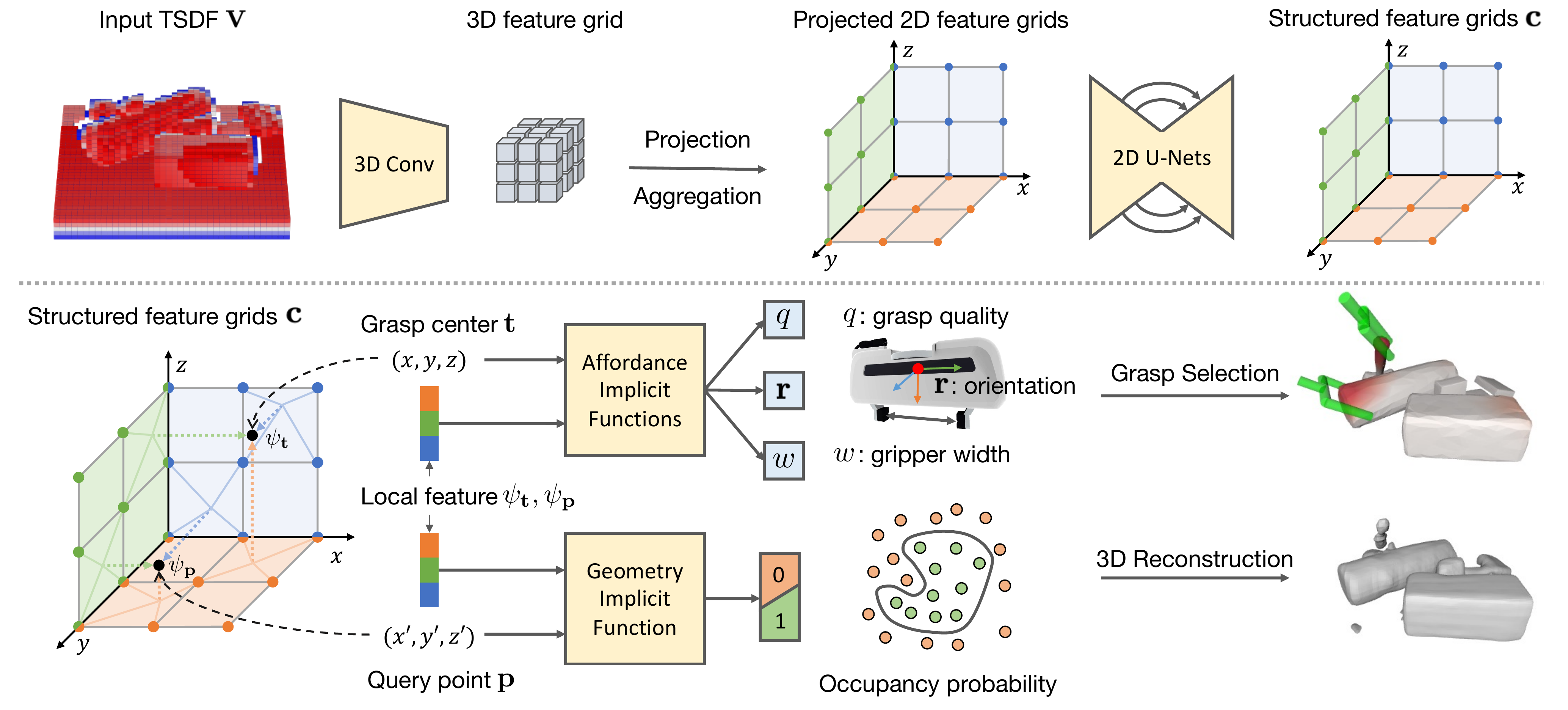}
    \caption{Model architecture of GIGA. The input is a TSDF fused from the depth image. After a 3D convolution layer, the output 3D voxel features are projected to canonical planes and aggregated into 2D feature grids. After passing each of the three feature planes through three independent U-Nets, we query the local feature at grasp center/occupancy query point with bilinear interpolation. The affordance implicit functions predict grasp parameters from the local feature at the grasp center. The geometry implicit function predicts occupancy probability from the local feature at the query point.} %
    \label{fig:arch}
\end{figure*}

\subsection{Implicit Neural Representations}

Implicit neural representations~\cite{sitzmann2020implicit} are continuous functions $\Phi$ that are parameterized by neural networks and satisfy equations of the form:
\begin{equation} \label{eqn:implicit-function}
    F(\mathbf{p}, \Phi_\theta)=0, \quad \Phi_\theta: \mathbf{p}\rightarrow \Phi_\theta(\mathbf{p})
\end{equation}
where $\mathbf{p}\in\mathbb{R}^m$ is a spatial or spatio-temporal coordinate, $\Phi_\theta$ is implicitly defined by relations defined by $F$, and $\theta$ refers to the parameters of the neural networks. As $\Phi_\theta$ is defined on continuous domains of $\mathbf{p}$ (\textit{e.g.}, Cartesian coordinates), it serves as a memory-efficient and differentiable representation of high-resolution 3D data compared to explicit and discrete representations.

\subsection{Occupancy Networks} \label{sec:onet-n-convonets}
Here we introduce Occupancy Network (ONet)~\cite{mescheder2019occupancy}, one of the pioneer works that learn implicit neural representations for 3D reconstruction.
Given an observation of a 3D object, ONet fits a continuous occupancy field defined over the bounding 3D volume of the object. The occupancy field is defined as:
\begin{equation}
o: \mathbb{R}^3 \rightarrow \{0, 1\}.
\end{equation}
Following Equation~\eqref{eqn:implicit-function}, the \textit{occupancy function} in ONet is defined as:
\begin{equation}
  \Phi_\theta(\mathbf{p}) - o(\mathbf{p}) = 0.
\end{equation}
The resulting occupancy function $\Phi_\theta(\mathbf{p})$ maps  a  3D location $\mathbf{p} \in \mathbb{R}^3$ to  the occupancy  probability between  0  and  1.

To generalize over shapes, we need to estimate occupancy functions conditioning on a context vector $x \in \mathcal{X}$ computed from visual observations of the shape. A function that conditions on the context $x \in \mathcal{X}$ and instantiates an implicit function $\Phi: \mathbb{R}^3 \rightarrow \mathbb{R}$ is equivalent to a function that takes a pair $(\mathbf{p}, x) \in \mathbb{R}^3 \times \mathcal{X}$ as input and has a real number as output. The latter function parameterized by a neural network $f_\theta$ is the Occupancy Network:
\begin{equation}
  f_\theta : \mathbb{R}^3 \times \mathcal{X} \rightarrow [0, 1].
\end{equation}

Our model builds on top of the architecture of Convolutional Occupancy Networks (ConvONets)~\cite{peng2020convolutional}, which is designed for scene-level 3D reconstruction. ConvONets extend ONet with a convolutional encoder and a structured representation. ConvONets encode input observation (point cloud or voxel grid) into structured feature grids, and condition implicit functions on local features retrieved from these feature grids. They demonstrate more accurate 3D reconstructions of local details in large-scale environments.

\section{Problem Formulation} \label{sec:problem-formulation}

We consider the problem of 6-DoF grasp detection for unknown rigid objects in clutter from a single-view depth image.

\subsection{Assumptions}

We assume a robot arm equipped with a parallel-jaw gripper in a cubic workspace with a planar tabletop. We initialize the workspace by placing multiple rigid objects on the tabletop. A single-view depth image taken with a fixed side view depth camera is fused into a Truncated Signed Distance Function (TSDF) and then passed into the model. The model outputs 6-DoF grasp pose predictions and associated grasp quality. We train our model with grasp trials generated in a self-supervised fashion from a physics engine. In simulated training, we assume the pose and shape of the objects are known.

\subsection{Notations}

\textbf{Observations}~ Given the depth image captured by the depth camera with known intrinsic and extrinsic, we fuse it into a TSDF, which is an $N^3$ voxel grid $\mathbf{V}$ where each cell $V_i$ contains the truncated signed distance to the nearest surface. We believe this additional distance-to-surface information provided by TSDF can improve grasp detection performance. Therefore we convert the raw depth image into TSDF volume $\mathbf{V}$ and pass it into the model.

\textbf{Grasps}~ We define a 6-DoF grasp $g$ as the grasp center position $\mathbf{t} \in \mathbb{R}^3$, the orientation $\mathbf{r} \in SO(3)$ of the gripper, and the opening width $w \in \mathbb{R}$ between the fingers.

\textbf{Grasp Quality}~ A scalar grasp quality $q\in[0,1]$ estimates the probability of grasp success. We learn to predict the grasp quality of a grasp with binary success labels of executing the grasp trial in simulation.%

\textbf{Occupancy}~ For an arbitrary point $\mathbf{p} \in \mathbb{R}^3 $, the occupancy $b \in \{0, 1\}$ is a binary value indicating whether this point is occupied by any of the objects in the scene.

\subsection{Objectives}
\vspace{1mm}

The primary goal is to detect 6-DoF grasp configurations that allow the robot arm to successfully grasp and remove the objects from the workspace. To foster multi-task learning of geometry reasoning and grasp affordance, we perform simultaneous 3D reconstruction. Given the input observation $\mathbf{V}$, our goal is to learn two functions:
\begin{equation}
    \begin{aligned}
        & f_a:\mathbf{t} \rightarrow q, \mathbf{r}, w, \\
        & f_g:\mathbf{p} \rightarrow b.
    \end{aligned}
\end{equation}
The first function $f_a$ maps from a grasp center to the rotation, gripper width, and grasp quality of the best grasp at that location. Once $f_a$ is trained, we can select which grasp to execute based on the grasp quality at a set of grasp centers. The second function $f_g$ maps any point in the workspace to the estimated occupancy value at that point. We can extract a 3D mesh from the learned occupancy function with the Marching Cube algorithm~\cite{lorensen1987marching}.%

\vspace{2mm}
\section{Method}
\vspace{2mm}

We now present GIGA, a learning framework that exploits synergies between affordance and geometry for 6-DoF grasp detection from partial observation. We learn grasp affordance prediction and 3D occupancy prediction jointly with shared feature grids and a unified implicit neural representation. Figure \ref{fig:arch} illustrates the overall model architecture.

\vspace{2mm}
\subsection{Structured Feature Grids}
\vspace{2mm}

To jointly learn the grasp affordance prediction and 3D reconstruction, we need to extract a shared feature from the input TSDF. In previous implicit-function-based 3D reconstruction works~\cite{mescheder2019occupancy,park2019deepsdf,van2020learning}, a flat global feature is extracted from the input observation and it is used to infer the implicit function. However, this simple representation falls short of encoding local spatial information, nor for incorporating inductive biases such as translational equivariance into the model~\cite{peng2020convolutional}. For this reason, they lead to overly smooth surface reconstructions. For grasp detection, local geometry reasoning is even more important. The spatial distribution of grasp affordance is very sparse and highly localized: most viable grasps cluster around a few graspable regions. Therefore, we adopt the encoder architecture from ConvONets~\cite{peng2020convolutional} and learn to extract structured feature grids from partial observation. 

Our encoder takes as input a TSDF voxel field and processes it with a 3D CNN layer to obtain a feature embedding for every voxel. Given these features, we construct planar feature representations by performing an orthographic projection onto a canonical plane for each input voxel. The canonical plane is discretized into pixel cells. Then we aggregate the features of voxels projected onto the same pixel cell using average pooling, which gives us a feature plane. The projection operation greatly reduces the computation cost while keeping the spatial distribution of feature points. We apply this feature projection and aggregation process to all three canonical frames. The feature plane might have empty features due to the incomplete observation. We therefore process each of these feature planes with a 2D U-Net~\cite{ronneberger2015u} which is composed of a series of down-sampling and up-sampling convolutions with skip connections. The U-Net integrates both local and global information and acts as a feature inpainting network. The output feature grids denoted as $\mathbf{c}$, are shared for affordance and geometry learning.

\subsection{Implicit Neural Representations} \label{sec:neural-implict-functions}

A unified representation of affordance and geometry facilitates the joint learning of grasp detection and 3D occupancy prediction. Recent works in the 3D vision community have shown that deep implicit functions are capable of representing shapes in a continuous and memory-efficient way and they can adaptively allocate computational and memory resources to regions of importance~\cite{genova2020local,liu2020neural,peng2020convolutional}. Therefore, we are motivated to use implicit neural representations to encode both affordance and geometry. As both require reasoning about local geometry details, we condition the deep implicit functions on local features. We query the local feature from the shared feature planes $\mathbf{c}$. Given a query position $\mathbf{p}$, we project it to each feature plane and query the features at the projected locations using bilinear interpolation. Unlike ConvONets~\cite{peng2020convolutional} where features from different planes are summed together, we concatenate them instead in order to preserve the spatial information. The local feature $\psi_\mathbf{p}$ can be formulated as:
\begin{equation}
    \psi_\mathbf{p} = [\phi (\mathbf{c}_{xy}, \mathbf{p}_{xy}), \phi (\mathbf{c}_{yz}, \mathbf{p}_{yz}), \phi (\mathbf{c}_{xz}, \mathbf{p}_{xz})],
\end{equation}
where $\mathbf{c}_{ij}, \mathbf{p}_{ij}~(i,j\in \{x, y, z\})$ are the plane feature and point projected onto the corresponding plane, and $\phi$ means bilinear interpolation of feature plane at the projected point.

\vspace{1mm}
\paragraph{Affordance Implicit Functions} The affordance implicit functions represent the grasp affordance field of grasp parameters and grasp quality. They map the grasp center $\mathbf{t}$ to grasp parameters ($\mathbf{r}$ and $w$) and the grasp quality metric $q$. These implicit neural representations enable learning directly from data with continuous grasp centers. In contrast, VGN has to snap grasp centers to the nearest voxel as it uses an explicit voxel-based grasp field. The snapping operation leads to information loss while our model does not.

We implement implicit neural representations as functions that map a pair of point and corresponding local feature $(\mathbf{t}, \psi_\mathbf{t})$ to target values. We parametrize the functions with small fully-connected occupancy networks with multiple ResNet blocks~\cite{he2016deep}. Grasp center, grasp orientation, and gripper width are output from three separate implicit functions: 
\begin{equation}
    \begin{aligned}
        f_{\theta_q} (\mathbf{t}, \psi_\mathbf{t}) &\rightarrow [0,1], \\
        f_{\theta_r} (\mathbf{t}, \psi_\mathbf{t}) &\rightarrow SO(3), \\
        f_{\theta_w} (\mathbf{t}, \psi_\mathbf{t}) &\rightarrow [0, w_{max}]. \\
    \end{aligned}
\end{equation}
Here $w_{max}$ is the maximum gripper width. $\theta_q, \theta_r, \theta_w$ are neural networks parameters of each deep implicit function. We represent grasp orientation with quaternions.

\vspace{1mm}
\paragraph{Geometry Implicit Function}
Our geometry implicit function maps from an arbitrary query point $\mathbf{p}$ inside the bounded volume to the occupancy probability $o(\mathbf{p})$ at the point. Similar to affordance implicit functions, we learn a function that predicts occupancy based on input point coordinate and the corresponding local feature:
\begin{equation}
    f_{\theta_o} (\mathbf{p}, \psi_\mathbf{p}) \rightarrow [0,1].
\end{equation}
Notice that the query points of occupancy $\mathbf{p}$ can be different from the grasp center points $\mathbf{t}$.

\subsection{Grasp Detection} \label{sec:grasp-detection}

GIGA takes as input a TSDF voxel grid, a grasp center, and multiple occupancy query points and predicts grasp parameters corresponding to the grasp center and occupancy probabilities at the query points. 

Given the trained GIGA model, we use a sampling procedure to select the final grasp pose. Grasp affordance is implicitly defined by the learned neural networks, so we need to query it from the learned implicit functions. To cover all possible graspable regions, we discretize the volume of the workspace into voxel grids and use the position of all the voxel cells as grasp centers. Then we query the grasp quality and grasp parameters corresponding to these grasp centers in parallel. Next, we mask out impractical grasps and apply non-maxima suppression as done in VGN~\cite{breyer2020volumetric}. Finally, we select a grasp with the highest quality if the quality is beyond a threshold. If no grasp has the quality above the threshold, we don't make grasp predictions and give up the current scene.

\subsection{Training}

The loss for training consists of two parts: the affordance loss and the geometry loss. For the affordance loss, we adopt the same training objective as VGN~\cite{breyer2020volumetric}:
\begin{equation}
    \centering
    \mathcal{L}_A(\hat{g}, g) = \mathcal{L}_q(\hat{q}, q) + q(\mathcal{L}_r(\hat{\mathbf{r}}, \mathbf{r}) + \mathcal{L}_w(\hat{w}, w)).
\end{equation}
Here $\hat{g}$ denotes predicted grasp parameters and $g$ denotes ground-truth parameters. $q\in\{0, 1\}$ is the ground-truth grasp quality (0 for failure, 1 for success) and $\hat{q}\in[0,1]$ is the predicted grasp quality. $\mathcal{L}_q$ is a binary cross-entropy loss between the predicted and ground-truth grasp quality. $\mathcal{L}_w$ is the $\ell_2$-distance between predicted gripper width $\hat{w}$ and ground-truth one $w$. For orientation, $\mathcal{L}_{quat}$ between predicted quaternion $\hat{\mathbf{r}}$ and target quaternion $\mathbf{r}$ is given by $\mathcal{L}_{quat}(\hat{\mathbf{r}}, \mathbf{r}) = 1 - |\hat{\mathbf{r}}\cdot \mathbf{r}|$. However, the parallel-jaw gripper is symmetric, which means a grasp configuration corresponds to itself after rotated by $180^\circ$ about the gripper's wrist axis. To handle this symmetry during training, both mirrored rotations $\mathbf{r}$ and $\mathbf{r}_\pi$ are deemed as ground-truth. Thus the orientation loss is defined as:
\begin{equation}
    \label{eqn:aff-loss}
    \mathcal{L}_r(\hat{\mathbf{r}}, \mathbf{r}) = \min (\mathcal{L}_{quat}(\hat{\mathbf{r}}, \mathbf{r}), \mathcal{L}_{quat}(\hat{\mathbf{r}}, \mathbf{r}_\pi)).
\end{equation}
We only supervise the grasp orientation and gripper width when a grasp is successful ($q=1$).

For the geometry loss, we apply the standard binary cross-entropy loss between the predicted occupancy $\hat{o}\in[0,1]$ and the ground-truth occupancy label $o\in\{0,1\}$. The loss is denoted as $\mathcal{L}_G$. The final loss is simply the direct sum of the affordance loss and geometry loss:
\begin{equation}
    \mathcal{L} = \mathcal{L}_A + \mathcal{L}_G.
\end{equation}
We implement the models with Pytorch~\cite{paszke2019pytorch} and train the models with the Adam optimizer~\cite{kingma2014adam} and a learning rate of $2\times 10^{-4}$ and batch sizes of 32.

\begin{figure}
    \captionsetup{font=footnotesize}
    \includegraphics[width=\linewidth]{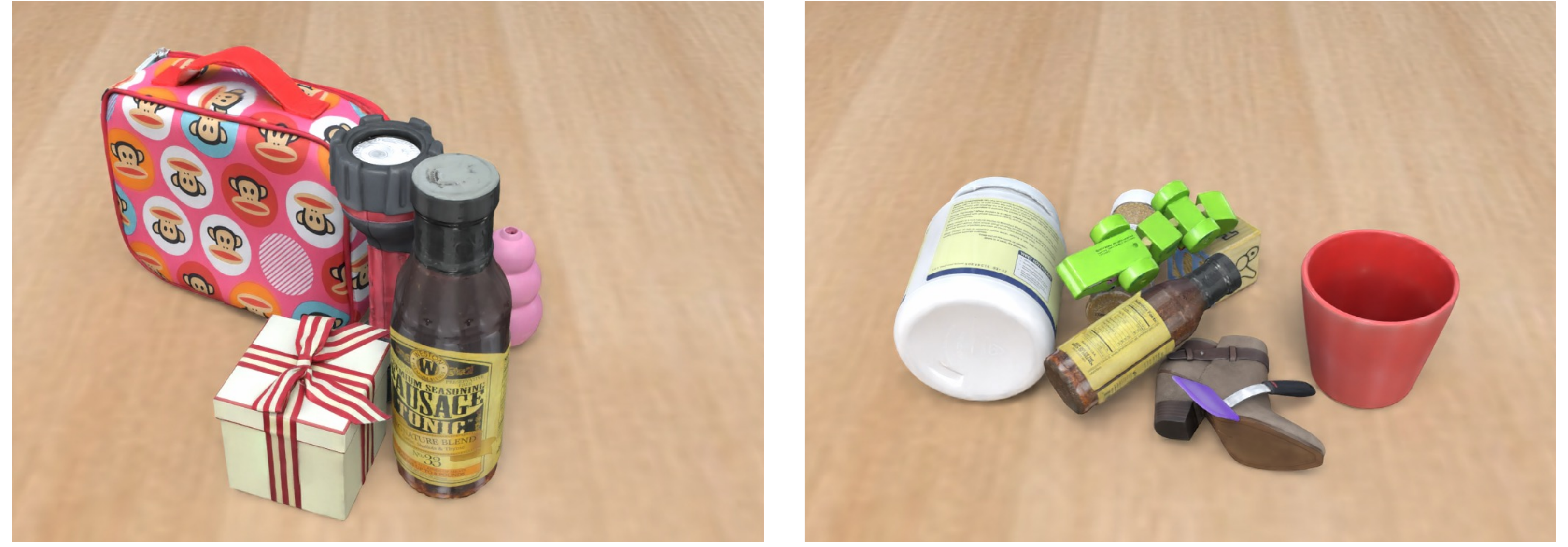}
    \caption{Visualization of packed (left) and pile (right) scenarios. In the packed scenario, objects are placed on the table at their canonical poses. In the pile scenario, objects are dropped on the workspace with random poses. These objects are from Google Scanned Objects~\cite{IgnitionFuel-GoogleResearch-Google-Scanned-Objects} and the scenes are rendered with NVISII~\cite{Morrical20nvisii}.}
    \label{fig:packed-n-pile}
\end{figure}

\section{Experiments}

We study the efficacy of synergies between affordance geometry on grasp detection in clutter. Specifically, we would like to investigate three complementary questions:
1) Can the structured implicit neural representations encode action-related information for grasping? 2) Does joint learning of geometry and affordance improve grasp detection? 3) How does grasp affordance learning impact the performance of 3D reconstruction?

\begin{table}[t]
    \captionsetup{font=footnotesize}
    \caption{Quantitative results of clutter removal. We report mean and standard deviation of grasp success rates (GSR) and declutter rates (DR). HR denotes high resolution.}
    \centering
    \label{tab:clutter-removel}
    \begin{tabular}{l c c c c}
        \toprule
        Method & \multicolumn{2}{c}{Packed} & \multicolumn{2}{c}{Pile} \\
        & GSR (\%) & DR (\%) & GSR (\%) & DR (\%) \\
        \midrule
        SHAF~\cite{fischinger2013learning} & $56.6 \pm 2.0$ & $58.0 \pm 3.0$ & $50.7 \pm 1.7$ & $42.6 \pm 2.8$ \\
        GPD~\cite{gualtieri2016high} & $35.4 \pm 1.9$ & $30.7 \pm 2.0$ & $17.7 \pm 2.3$ & $9.2 \pm 1.3$ \\
        VGN~\cite{breyer2020volumetric} & $74.5 \pm 1.3$ & $79.2 \pm 2.3$ & $60.7 \pm 4.2$ & $44.0 \pm 4.9$ \\
        \midrule
        GIGA-Aff & $77.2 \pm 2.3$ & $78.9 \pm 1.7$ & $67.8 \pm 3.0$ & $49.7 \pm 1.9$ \\
        GIGA & $83.5 \pm 2.4$ & $84.3 \pm 2.2$ & $69.3 \pm 3.3$ & $49.8 \pm 3.9$ \\
        GIGA (HR) & $\mathbf{87.9 \pm 3.0}$& $\mathbf{86.0 \pm 3.2}$ & $\mathbf{69.8 \pm 3.2}$& $\mathbf{51.1 \pm 2.8}$ \\
        \bottomrule
    \end{tabular}
\end{table}

\subsection{Experimental Setup}
Our model is trained in a self-supervised manner with ground-truth grasp labels collected from physical trials in simulation and occupancy data obtained from the object meshes. The use of TSDF enables zero-shot transfer of our model from simulation to a real Panda arm from Franka Emika.

\vspace{1mm}
\paragraph{Simulation Environment} Our simulated environment is built on PyBullet~\cite{coumans2019}. We use a free gripper to sample grasps in a $30 \times 30 \times 30 \text{cm}^3$ tabletop workspace. For a fair comparison, we use the same object assets as VGN, including 303 training and 40 test objects from different sources~\cite{calli2015ycb,kappler2015leveraging,kasper2012kit,singh2014bigbird}. The simulation grasp evaluations are all done with the test objects, which are excluded from training.

We collect grasp data in a self-supervised fashion in two type of simulated scenes, \textit{pile} and \textit{packed} as in VGN~\cite{breyer2020volumetric}. In the pile scenario, objects are randomly dropped to a box of the same size as the workspace. Removing the box leaves a cluttered pile of objects. In the packed scenario, a subset of taller objects is placed at random locations on the table at their canonical pose. Examples of these two scenarios are shown in Figure~\ref{fig:packed-n-pile}. Once the scene is created, we randomly sample grasp centers and grasp orientations near the surface of the objects and execute these grasp samples in simulation. We store grasp parameters and the corresponding outcomes of the grasp trials and balance the dataset by discarding redundant negative samples.

We collect the occupancy training data in the same scenes where grasp trials are performed. Upon the creation of a simulation scene, we query the binary occupancy of a large number of points uniformly distributed in the cubic workspace as the training data.

\vspace{1mm}
\paragraph{Camera Observations} %
To evaluate the model's robustness against noise and occlusion in real-world clutter, we assume that the robot perceives the workspace by a single depth image from a fixed side view. If we denote the length of the workspace as $l$ and use the spherical coordinates with the workspace center as the origin, the viewpoint of the virtual camera points towards the origin at $r=2l,\theta=\frac{\pi}{3},\phi=0$. To expedite sim-to-real transfer, we add noise to the rendered images in simulation using the additive noise model~\cite{mahler2017dex}:
\begin{equation}
    \mathbf{y}=\alpha\hat{\mathbf{y}} + \epsilon,
\end{equation}
where $\hat{\mathbf{y}}$ is a rendered depth image, $\alpha$ is a Gamma random variable with $k=1000$ and $\theta=0.001$ and $\epsilon$ is a Gaussian Process noise drawn with measurement noise $\sigma=0.005$ and kernel bandwidth $l=\sqrt{2}px$. The input to the our algorithm is a $40\times 40 \times 40$ TSDF~\cite{curless1996volumetric} fused from this noisy single-view depth image using the Open3D library~\cite{Zhou2018}.

\vspace{1mm}
\paragraph{Grasp Execution} We select top grasps to execute by querying grasp parameters from the learned implicit functions with a set of grasp centers. For a fair comparison with VGN~\cite{breyer2020volumetric}, our \textbf{GIGA} model samples $ 40\times 40 \times 40$ uniformly distributed grasp centers in the workspace and query the grasp parameters. However, our implicit representations are continuous, so we can query grasp samples in arbitrary resolutions. In \textbf{GIGA (HR)}, we query at a higher resolution of $60 \times 60 \times 60$.

We use a set of clutter removal scenarios to evaluate GIGA and other baselines. Each round, a pile or packed scene with 5 objects is generated. We take a depth image from the same viewpoint as training. The grasp detection algorithm generates a grasp proposal given the input TSDF. We execute the grasp and remove the grasped object from the workspace. If all objects are cleared, two consecutive failures happen, or no grasp is detected, we terminate the current scene. Otherwise, we collect the new observation and predict the next grasp. In our experiments, grasp proposals with a predicted grasp quality below 0.5 are discarded.

\begin{figure}
    \captionsetup{font=footnotesize}
    \centering
    \includegraphics[width=\linewidth]{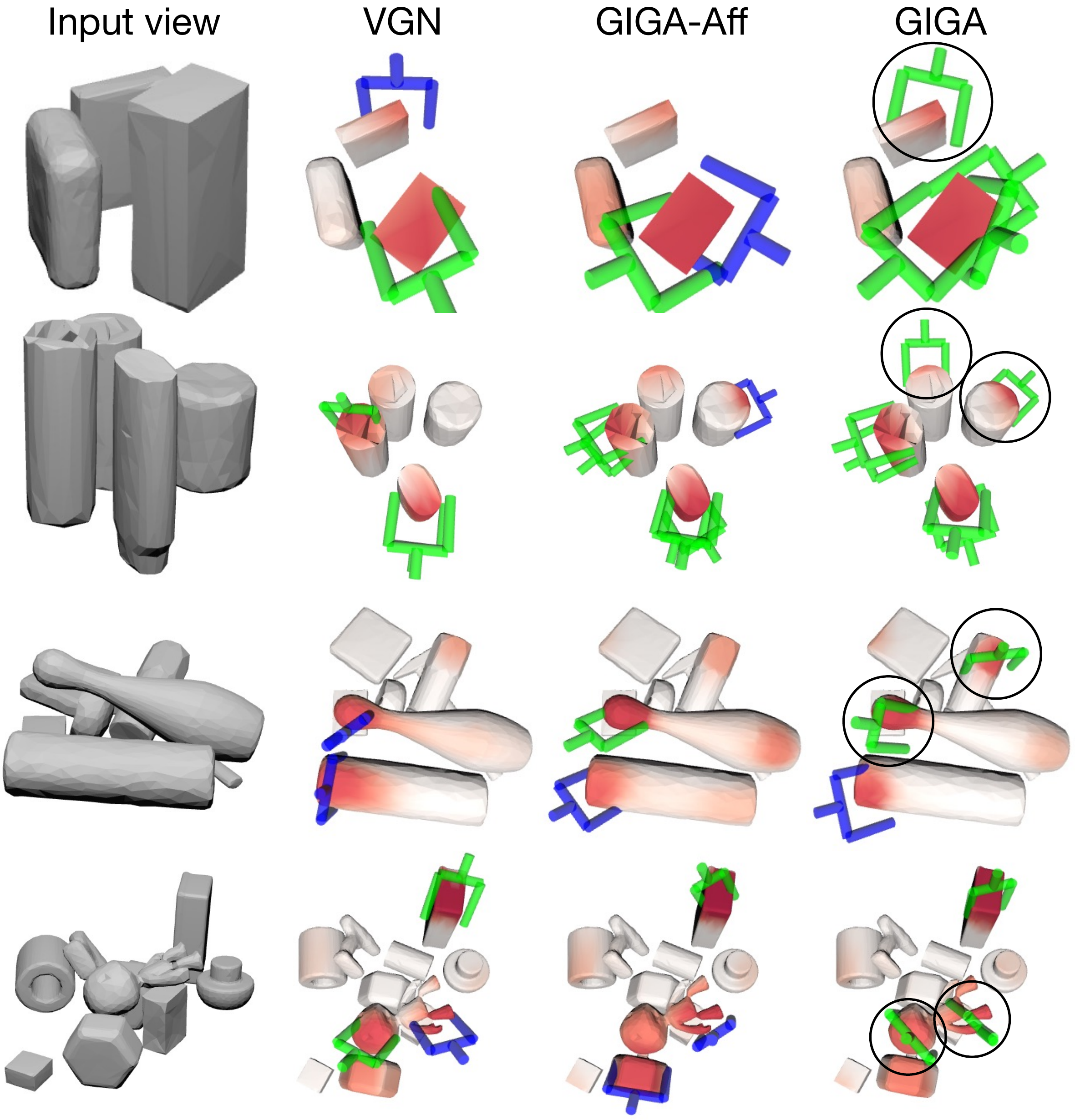}
    \caption{Visualization of the grasp affordance landscape and predicted grasps. \textcolor{red}{\textbf{Red}} indicates that the method predicts high grasp affordance near the corresponding area. \textcolor{ForestGreen}{\textbf{Green}} indicates successful grasps and \textcolor{Blue}{\textbf{Blue}} failures. The circles highlight interesting examples, such as asymmetric affordance heatmaps and highly occluded objects.}
    \label{fig:aff}
\end{figure}

\subsection{Baselines}

We compare the performance of our method and the following baselines:

\begin{itemize}
    \item \textbf{SHAF}: We use the highest point heuristic~\cite{fischinger2013learning} by classic work of grasping in clutter, rather than the learned grasp quality, for grasp selection. Among all grasps of quality over 0.5 predicted by our model, we select the highest one to grasp from the clutter.
    \item \textbf{GPD}~\cite{gualtieri2016high}: Grasp Pose Detection, a two-stage 6-DoF grasp detection algorithm that generates a large set of grasp candidates and classifies each of them.
    \item \textbf{VGN}~\cite{breyer2020volumetric}: Volumetric Grasping Network, a single-stage 6-DoF grasp detection algorithm that generates a large number of grasp parameters in parallel given input TSDF volume.
    \item \textbf{GIGA-Aff}: An ablated version of our method with only affordance implicit function branch. The network is trained with only grasp supervision but no reconstruction.
\end{itemize}

Performance is measured using the following metrics averaged over 100 simulation rounds: 1) Grasp success rate (GSR), the ratio of success grasp executions; and  2) Declutter rate (DR), the average ratio of objects removed. The original VGN uses multi-view inputs, we re-train the VGN model on the same single-view data we used for training GIGA for fair comparisons. %

\subsection{Grasp Detection Results}

We report grasp success rate and declutter rate for different scenarios in Table \ref{tab:clutter-removel}. %
We can see that GIGA and GIGA-Aff outperform other baselines in almost all scenarios and metrics. Even though GIGA-Aff does not utilize the synergies between affordance and geometry and is trained without geometry supervision, it still outperforms the state-of-the-art VGN baseline. We attribute this to the high expressiveness of our implicit neural representations. VGN snaps the grasp center to voxel grid cells during training. In contrast, our implicit neural representations learn to fit grasp affordance field with continuous functions. It allows us to query grasp parameters at a higher resolution as done in GIGA (HR). GIGA (HR) gives the highest performance in all cases.

\begin{figure}
    \captionsetup{font=footnotesize}
    \centering
    \includegraphics[width=\linewidth]{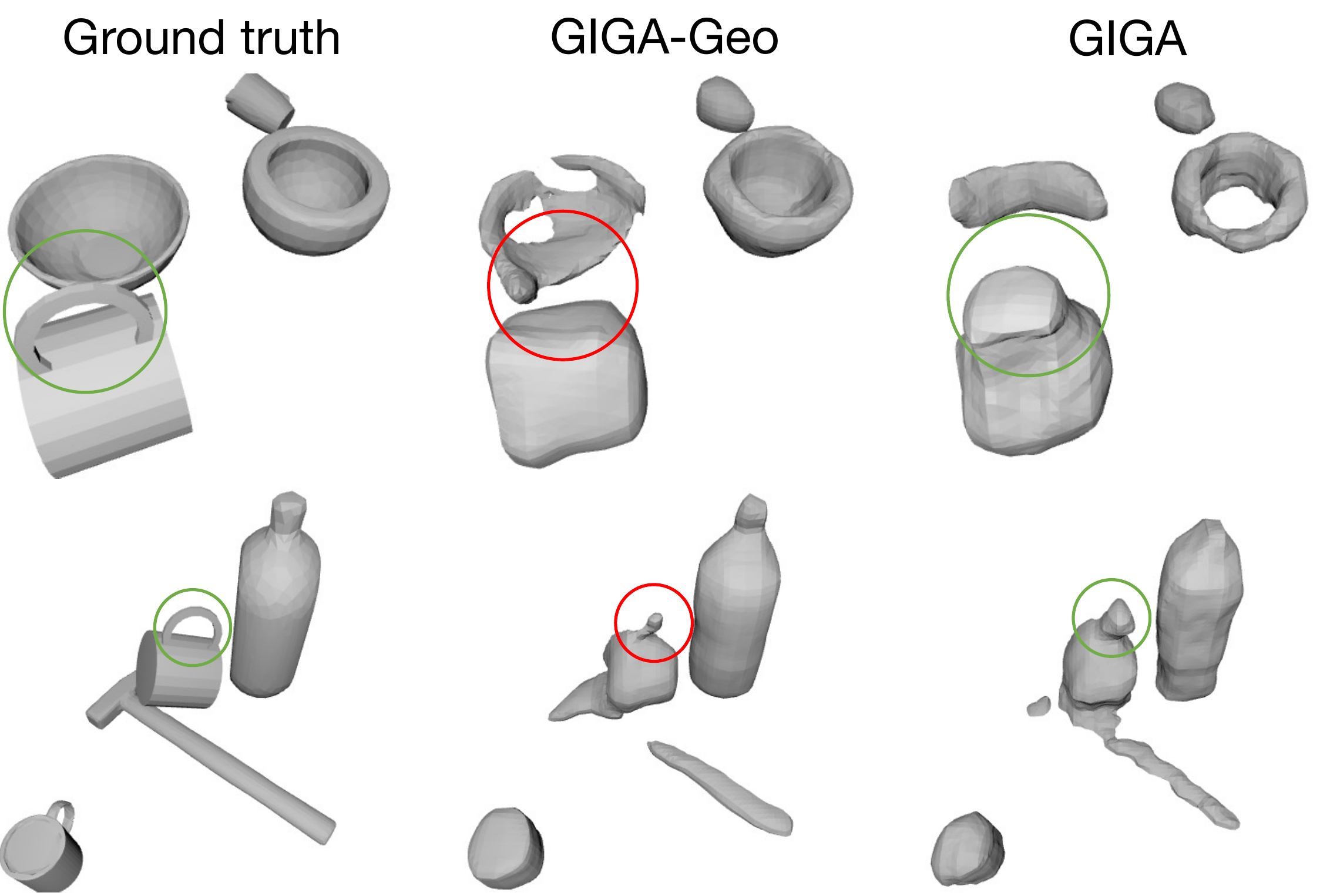}
    \caption{Qualitative 3D reconstruction results of a scene rendered from the top view. The circles highlight the contrast.}
    \label{fig:3D_recon}
\end{figure}

Next, we compare the results of GIGA-Aff with GIGA. In the pile scenario, the gain from geometry supervision is relatively small (around 2\% grasp success rate). However, in the packed scenario, GIGA outperforms GIGA-Aff by a large margin of around 5\%. We believe this is due to the different characteristics of these two scenarios. From Figure~\ref{fig:packed-n-pile}, we can see that in the packed scene, some tall objects standing in the workspace would occlude the objects behind them and the occluded objects are partially visible. We hypothesize that in this case, the geometrically-aware feature representation learned via geometry supervision facilitates the model to predict grasps on partially visible objects. Such occlusion is, however, less frequent in pile scenarios. These results demonstrate that synergies between affordance and geometry improve grasp detection, especially in the presence of occlusion.

We show examples of the learned grasp affordance landscape (grasp quality at different locations of the scene) and predicted top grasps in Figure~\ref{fig:aff}. Our first observation is that our method is able to take into account the context of the scene and predict collision-free grasps. It often predicts asymmetric affordance heatmap over symmetric objects (\textit{e.g.}, boxes and cylinders), where one part of the object is graspable (red) while the symmetric part is not (grey), and grasping these grey regions is likely to lead to a collision with the neighboring objects. This indicates that our model encodes scene-level information from training on self-supervised grasp trials and takes into account practical constraints when making grasp predictions.

Our second observation is that GIGA produces more diverse and accurate grasp detections compared with the baselines. In the first two rows, we visualize the affordance and grasp of two packed scenes. In regions marked by the black circle, GIGA predicts high affordance and more accurate grasps. The objects in these regions are largely occluded from the view but are easy to grasp given full 3D information. For example, the thin box in the circle of the first row affords a wide range of grasp poses. These easy grasps are not detected in GIGA-Aff or VGN due to occlusion but are successfully predicted by GIGA. The last two rows show the affordance landscape and top grasps for two pile scenes. We see that baselines without the multi-task training of 3D reconstruction tend to generate failed or no grasp, whereas GIGA produces more diverse and accurate grasps due to the learned geometrically-aware representations.

Another advantage of GIGA is the improvement of the grasping efficiency. The planning time(time between receiving depth image(s) and returning a list of grasps) of GIGA and original multi-view VGN is similar, which are 46ms(GIGA) and 46ms(VGN) on an NVIDIA Titan RTX 2080Ti. However, GIGA uses single-view depth input, which can be collected from a fixed camera instantaneously. Meanwhile, the original VGN model has to collect multi-view inputs by moving the wrist-mounted camera along a scanning trajectory. The estimated time cost of the scanning process before each grasp is about 16-20s. Therefore, GIGA grasps at least one order of magnitude faster than the VGN.

To further verify GIGA's ability to predict grasps from partial observation, we retrain GIGA on a fused TSDF of multi-view depth images taken from six randomly distributed viewpoints along a circle in the workspace, the same setup as VGN~\cite{breyer2020volumetric}. Compared to the single-view setup, acquiring multi-view observations is more time-consuming and subject to environmental constraints. Yet with multi-view inputs, GIGA achieves 88.8\% and 69.6\% grasp success rates in packed and pile scenarios, which is on par with the performances of GIGA on single-view input reported in Table \ref{tab:clutter-removel}. We hypothesize that supervision from 3D reconstruction plays a key role for GIGA to reason about the occluded parts of the scene.

\begin{table}[t]
    \captionsetup{font=footnotesize}
    \centering
    \caption{Quantitative results of 3D reconstruction. We see that models trained with grasp supervision tend to have better reconstruction performance near graspable regions than average by a larger margin.}
    \label{tab:3d-reconstruction}
    \begin{tabular}{l c c c}
        \toprule
        Method & IoU (\%) & IoU-Grasp (\%) & $\Delta\%$ (IoU-Grasp$-$IoU)\\
        \midrule
        GIGA-Detach & 53.2 & 68.8 & \textbf{+15.6} \\
        GIGA & 70.0 & 78.1 & +8.1 \\
        GIGA-Geo & \textbf{80.0} & \textbf{84.0} & +4.0\\
        \bottomrule
    \end{tabular}
\end{table}

\subsection{3D Reconstruction}

In the previous section, we have demonstrated that geometry supervision benefits grasp affordance learning. We now examine how grasp affordance learning affects 3D reconstruction. To this end, we evaluate the performance of 3D reconstruction on pile scenes. We further compare two ablated versions of our method: \textbf{GIGA-Geo}, which is trained solely for 3D reconstruction without the affordance implicit functions branch, and \textbf{GIGA-Detach}, which is trained to reconstruct the scene on the fixed feature grids from GIGA-Aff. %

We extract mesh from the learned implicit function through marching cube algorithm~\cite{lorensen1987marching}. We use the Volumetric IoU metric for evaluation, which is the intersection over the union between predicted mesh and ground-truth one. For comparison, we also report the IoU near graspable parts given by grasp trials, denoted as IoU-Grasp. Specifically, we sample the successful grasps and evaluate the IoU in the regions between the two fingers of these grasps.

Table \ref{tab:3d-reconstruction} shows the results of 3D reconstruction. GIGA-Geo gives the highest overall IoU and the highest IoU in graspable parts. This result is unsurprising since GIGA-Geo is trained to specialize in this task. In contrast, both GIGA and GIGA-Detach are trained to predict grasps, of which the spatial distribution is highly localized. Thus, we expect them to biased their representational resources towards these graspable regions.

\begin{table}[t]
    \captionsetup{font=footnotesize}
    \centering
    \caption{Quantitative results of clutter removal in real world. We report GSR, DR, the number of successful grasps, and the number of total grasp trials (in bracket).}
    \label{tab:clutter-removel-real}
    \begin{tabular}{l c c c c}
        \toprule
        Method & \multicolumn{2}{c}{Packed} & \multicolumn{2}{c}{Pile} \\
        & GSR (\%) & DR (\%) & GSR (\%) & DR (\%) \\
        \midrule
        VGN~\cite{breyer2020volumetric} & 77.2 (61\,/\,79) & 81.3 & 79.0 (64\,/\,81) & 85.3 \\
        GIGA & \textbf{83.3} (65\,/\,78) & \textbf{86.6} & \textbf{86.9} (73\,/\,84) & \textbf{97.3} \\
        \bottomrule
    \end{tabular}
\end{table}

\begin{figure}[t]
    \captionsetup{font=footnotesize}
    \centering
    \includegraphics[width=\linewidth]{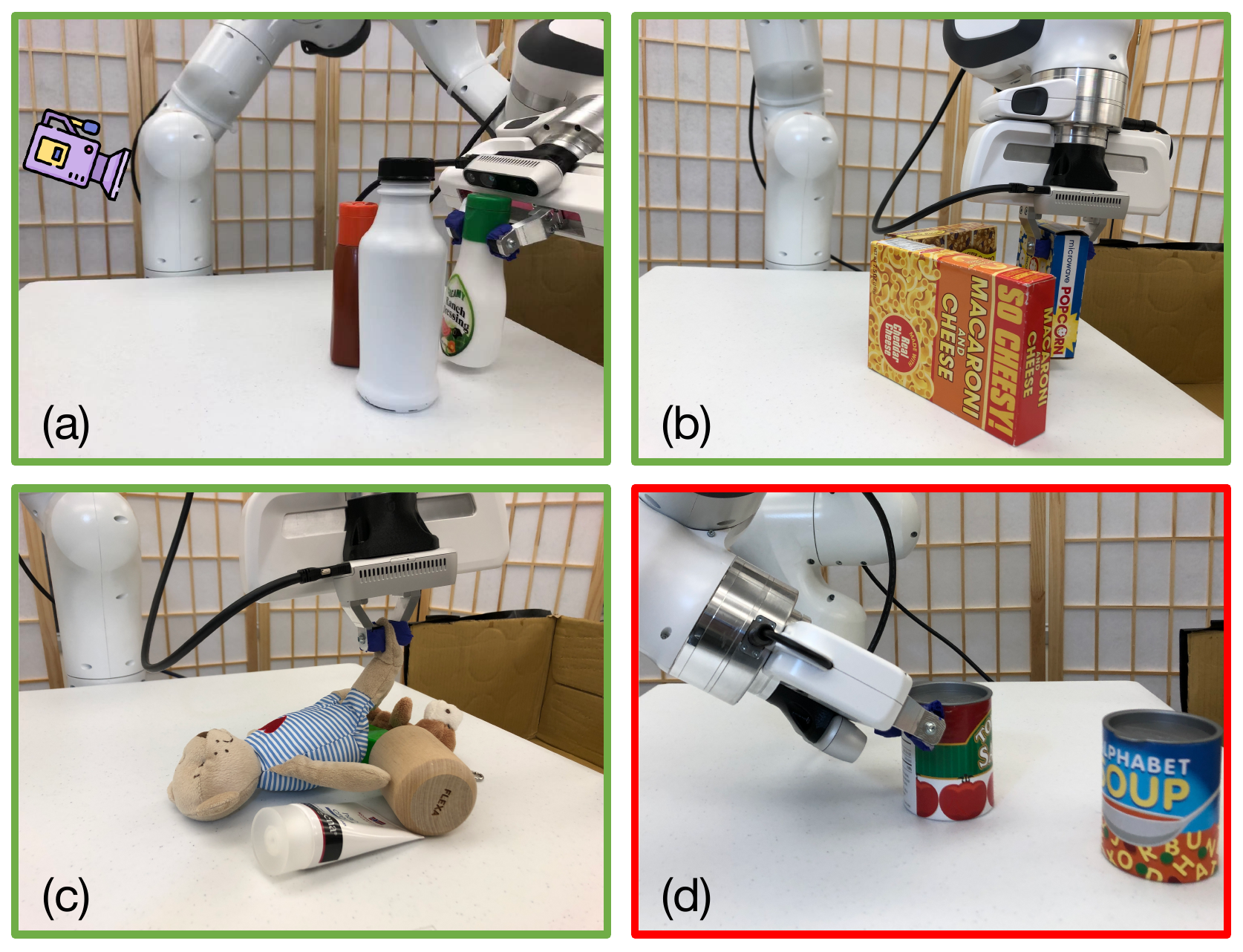}
    \caption{Examples of real-world grasps by GIGA. (a) and (b) show two examples of grasps of partially occluded objects. The tilted camera looks at the workspace from the left. (c) shows an example where GIGA picks up the bear doll from a localized graspable part. (d) illustrates a typical failure where the gripper slips off the object due to the small contact surface.}
    \label{fig:real_pic}
\end{figure}

Interesting observations arise when we look at the performance gaps between IoU-Grasp and IoU for the same method. GIGA-Detach gives the largest delta, then comes the GIGA. 
It indicates that scene features learned from grasp supervision dynamically allocate the representational budget to actionable regions. In Figure \ref{fig:3D_recon}, GIGA-Geo shows more accurate overall reconstruction results than GIGA, consistent with the quantitative results. In comparison, some parts are missing in the GIGA results, such as the bottom of the bowl. However, the missing parts are mostly non-graspable parts, while graspable parts such as the edges of the bowls are clearly reconstructed. In addition, we see that GIGA successfully reconstructs the handle of the cup, while GIGA-Geo either ignores the handle (in the first row) or reconstructs a small proportion of the handle. These results imply that GIGA attends to the action-related regions of high grasp affordance, driven by the grasp supervision.

\subsection{Real Robot Experiments}
Finally, we test our method in the clutter removal experiments on the real hardware. 15 rounds of experiments are performed with GIGA and VGN for both the Packed and Pile scenarios respectively. In each round, 5 objects are randomly selected and placed on the table. In each grasp trial, we pass the TSDF from a side view depth image to the model and execute the predicted top grasp. A grasp trial is marked as a success if the robot grasps the object and places it into a bin next to the workspace. We repeatedly plan and execute grasps till either all objects are cleared or two consecutive failures occur.

Table \ref{tab:clutter-removel-real} reports the real-world evaluations. GIGA achieves higher success rates and clears more objects. We show some grasp examples in Figure \ref{fig:real_pic}. Compared with VGN, GIGA is better at detecting grasps for partially occluded objects and finding graspable object parts such as edges and handles. We also show a failure case. In Figure \ref{fig:real_pic}(d), the grasp fails due to the insufficient contact surface. We believe this is attributed to the unrealistic contact and friction models in the simulated environments that we used for generating training data. Videos of clutter removal can be found in the supplementary video and on our website.

\section{Conclusion}

We introduced GIGA, a method for 6-DoF grasp detection in a clutter removal task. Our model learns grasp detection and 3D reconstruction simultaneously using implicit neural representations, exploiting the synergies between affordance and geometry. Concretely, we represent the grasp affordance and occupancy field of an input scene with continuous implicit functions, parametrized with neural networks. We learn shared structured feature grids on multi-task grasp and 3D supervision. In experiments, we study the influence of geometry supervision on grasp affordance learning and vice versa. The results demonstrate that utilizing the synergies between affordance and geometry can improve 6-DoF grasp detection, especially in the case of large occlusion, and grasp supervision produces better reconstruction in graspable regions.

Our method can be extended and improved in several fronts. First, in the training process, we assume only a single ground-truth grasp pose is provided for each grasp center. We hope to extend the current work and learn to predict the full distribution of viable grasp parameters with generative modeling. Second, though our model implicitly learns to avoid collisions from the training data where collided grasps are marked as negative, we do not explicitly reason about collision-free paths to grasps. We plan to utilize the reconstructed 3D scene to constrain the grasp prediction to be collision-free. We also plan to adapt this learning paradigm to close-loop grasp planning by integrating real-time feedback. %

\section*{Acknowledgments}
We would like to thank Zhiyao Bao for efforts on affordance visualization. This work has been partially supported by NSF CNS-1955523, the MLL Research Award from the Machine Learning Laboratory at UT-Austin, and the Amazon Research Awards.

\bibliographystyle{plainnat}
\bibliography{references}

\end{document}